\documentclass[sigconf]{acmart}

\AtBeginDocument{%
}

\copyrightyear{}
\acmYear{2026}
\setcopyright{cc}
\setcctype{by}
\acmConference[KDD '26]{Proceedings of the 32nd ACM SIGKDD Conference on Knowledge Discovery and Data Mining V.2}{August 09--13, 2026}{Jeju Island, Republic of Korea}
\acmBooktitle{Proceedings of the 32nd ACM SIGKDD Conference on Knowledge Discovery and Data Mining V.2 (KDD '26), August 09--13, 2026, Jeju Island, Republic of Korea}
\acmDOI{10.1145/3770855.3817848}
\acmISBN{979-8-4007-2259-2/2026/08}

\usepackage{amsmath}
\usepackage{amsthm}
\newtheorem{theorem}{Theorem}

\newtheorem{definition}{Definition}

\usepackage{diagbox}
\usepackage{graphicx}
\usepackage{algorithm}
\usepackage{algorithmic}
\usepackage{multirow}
\usepackage{xcolor}
\usepackage{array}
\usepackage{multirow}
\usepackage{booktabs}
\usepackage{amsfonts}
\usepackage{booktabs}
\usepackage{threeparttable}
\usepackage{newfloat}
\usepackage{listings}

\begin{document}

\title{ Perturbation Effects on Robustness and Individual Fairness}

\author{Xuran Li}
\orcid{0009-0008-3485-2603}
\affiliation{
  \institution{University of New South Wales}
  \city{Sydney}
  \country{Australia}
}
\email{xuran.li1@unsw.edu.au}

\author{Hao Xue}
\orcid{0000-0003-1700-9215}
\affiliation{%
  \institution{The Hong Kong University of Science and Technology}
  \city{Guangzhou}
  \country{China}
}
\email{haoxue@hkust-gz.edu.cn}

\author{Peng Wu}
\orcid{0000-0002-4931-0566}
\affiliation{%
  \institution{Key Laboratory of System Software, Institute of Software, Chinese Academy of Sciences}
  \city{Beijing}
  \country{China}
}
\email{wp@ios.ac.cn}

\author{Xingjun Ma}
\orcid{0000-0003-2099-4973}
\affiliation{%
  \institution{Fudan University}
  \city{Shanghai}
  \country{China}
}
\email{xingjunma@fudan.edu.cn}

\author{Zhen Zhang}
\orcid{0000-0001-7547-4313}
\affiliation{%
  \institution{Hangzhou Institute for Advanced Study, University of Chinese Academy of Sciences}
  \city{Hangzhou}
  \country{China}
}
\email{zhangzhen19@ios.ac.cn}

\author{Huaming Chen}
\orcid{0000-0001-5678-472X}
\affiliation{%
  \institution{The University of Sydney}
  \city{Sydney}
  \country{Australia}
}
\email{huaming.chen@sydney.edu.au}

\author{Flora D. Salim}
\orcid{0000-0002-1237-1664}
\affiliation{%
  \institution{University of New South Wales}
  \city{Sydney}
  \country{Australia}
}
\email{flora.salim@unsw.edu.au}

\renewcommand{\shortauthors}{Xuran Li et al.}

\begin{abstract}
Deep neural networks are vulnerable to adversarial perturbations that can simultaneously degrade prediction robustness and individual fairness across diverse application settings. However, existing evaluation protocols typically assess these dimensions in isolation, thereby obscuring critical failure modes.
To bridge this gap, we formalize \emph{Robust Individual Fairness} (\textsc{RIF}): under \emph{semantic-preserving} (truth-condition-preserving) perturbations, predictions should remain both correct with respect to the ground truth and invariant across semantically equivalent individuals. To surface \textsc{RIF} violations in practice, we introduce \textsc{RIFair}, a black-box adversarial framework that leverages a decoupled perturbation strategy to construct semantically preserved yet unrobust and/or unfair instance pairs.
Experiments across multiple model architectures and real-world textual datasets show that robustness-only or fairness-only metrics often miss \emph{Robust Biased} and \emph{Unrobust Fair} behaviors. \textsc{RIFair} reliably exposes these hidden vulnerabilities, supporting \textsc{RIF} as a necessary criterion for trustworthy model assessment. 
The experimental code is publicly available at \url{https://github.com/Xuran-LI/RIFair}.

\end{abstract}

\ccsdesc[500]{Computing methodologies~Machine learning}
\ccsdesc[500]{Security and privacy~Formal methods and theory of security}
\ccsdesc[500]{Social and professional topics~User characteristics}


\keywords{Trustworthy AI;
Adversarial attacks and perturbations;
Fairness--robustness interactions}
\maketitle

\section{Introduction}

\paragraph{Motivation.} Deep neural networks (DNNs) deliver strong predictive performance across domains, yet their deployment raises persistent concerns about trustworthiness~\cite{ zhou2026boosting, li2026safetyembodiedaisurvey,10.1145/3677173,10.1145/3450268.3453519}. Prior work shows that DNNs can be highly sensitive to minor, often imperceptible perturbations~\cite{test_survey,trustworthiness_1, FGSM}, which may simultaneously \textbf{(1)} compromise prediction robustness~\cite{FGSM,Adv2} and \textbf{(2)} undermine individual fairness through biased decisions~\cite{ADF,individually-fairness1}. Ensuring that model behavior remains both reliable and equitable under perturbations is therefore essential in many real-world settings.

\begin{figure}[ht]
    \centering
    \resizebox{1\columnwidth}{!}{
    \includegraphics[width=1.\textwidth]{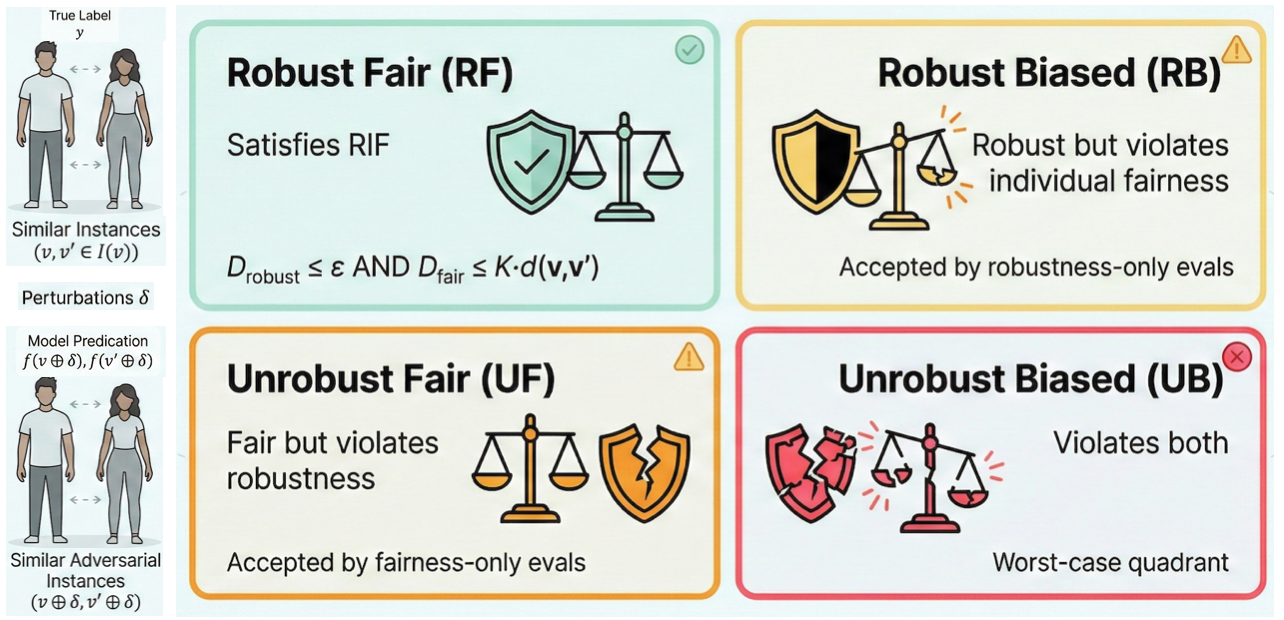}
    }
    \caption{The Robustness--Fairness Relationship.}
    \label{fig:R F Relationship}
\end{figure}

\begin{figure*}[t]
    \centering
    \includegraphics[width=1.\linewidth]{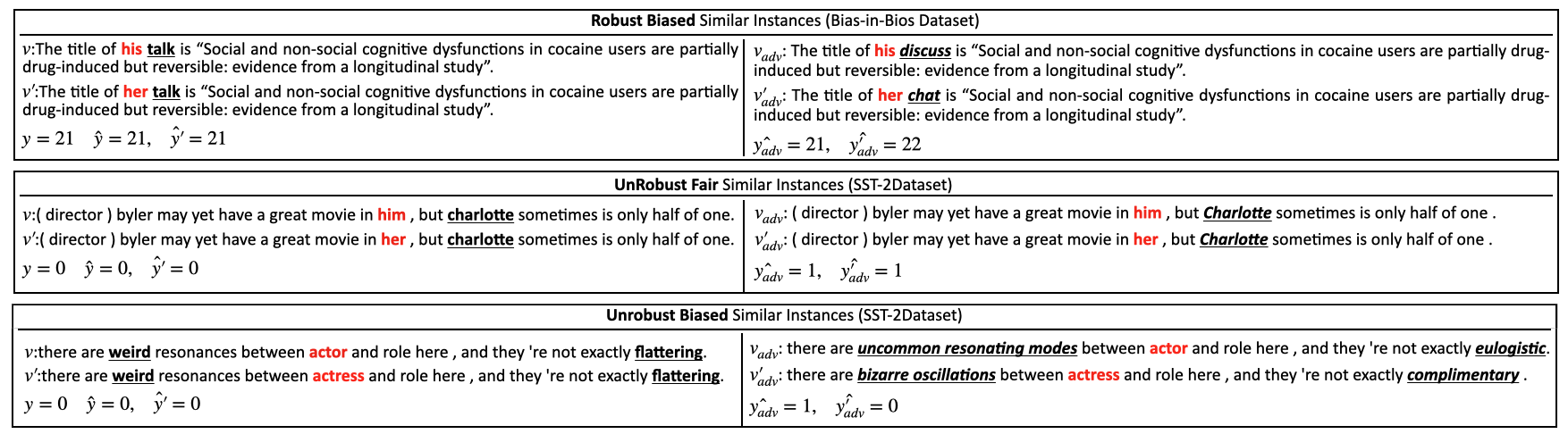}
    \caption{UnRobust or Unfair Adversarial Similar Instances by RIFair}
    \label{fig:UnRobust_or_Unfair}
\end{figure*}

\paragraph{Problem: robustness–fairness misalignment.}
Existing evaluation protocols often treat robustness and fairness as separate objectives:
attacks are either optimized for misclassification~\cite{FGSM} or designed to
surface discriminatory behavior ~\cite{ADF,RobustnessBias}. Consequently, single-axis metrics can obscure coupled failure modes. Figure~\ref{fig:R F Relationship} highlights that robustness-only and fairness-only metrics may overlook \emph{Robust Biased} and \emph{Unrobust Fair} behaviors. As a result, robustness resistance ($R$)~\cite{FGSM,textfooler} and individual-fairness resistance ($F$) may be inconsistent, and models can receive conflicting rankings depending on the evaluation axis (Table~\ref{tab:attack_metrics}). This misalignment is especially critical in high-stakes applications---including toxic content identification~\cite{jigsaw}, profession identification~\cite{Bias_in_Bios}, sentiment analysis~\cite{SST-2}, and income prediction~\cite{adult}---where both reliability and equity are essential.

\begin{table}[htbp]
\centering
\caption{Trustworthiness Rankings under Perturbations}
\label{tab:attack_metrics}
\resizebox{1.0\columnwidth}{!}{
\begin{tabular}{l|cc|cc|cc}
\hline
\multirow{2}{*}{{Model}} & \multicolumn{2}{c|}{{RIF}} & \multicolumn{2}{c|}{{Robustness}} & \multicolumn{2}{c}{{Fairness}} \\ 
 & $RIF$ & {Rank} & {$R$} & {Rank} & {$F$} & {Rank} \\ \hline
\textbf{DeBERTa-v3} & \textbf{0.605} & \textbf{1} & 0.704 & 3 & 0.972 & 4 \\
BERT & 0.592 & 2 & 0.713 & 2 & \textbf{0.977} & \textbf{1} \\
RoBERTa & 0.580 & 3 & 0.700 & 4 & 0.975 & 2 \\
DistilBERT & 0.574 & 4 & \textbf{0.717} & \textbf{1} & 0.972 & 3 \\ \bottomrule
\end{tabular}%
}
\end{table}

\paragraph{Proposed criterion: Robust Individual Fairness.}
To address this misalignment, we propose \emph{Robust Individual Fairness} (\textsc{RIF}), a unified criterion grounded in truth-conditional semantics~\cite{Truth-Conditional_Semantics} that targets the \emph{Robust Fair} regime in Figure~\ref{fig:R F Relationship}. \textsc{RIF} formalize trustworthiness as a joint requirement: for semantically equivalent (truth-condition-preserving) individuals, a model should (i) remain correct with respect to the ground truth and (ii) produce consistent outcomes across the similar individuals under admissible perturbations.
Beyond evaluation, \textsc{RIF} reflects a broader trustworthiness principle: \emph{a system should not yield materially different outcomes for inputs that are equivalent in task-relevant content, particularly when discrepancies stem from sensitive attributes or superficial surface-form variations rather than substantive semantic changes.}

\paragraph{Method: \textsc{RIFair}.}
To operationalize \textsc{RIF} and surface RB/UB/UF violations, we introduce \textsc{RIFair}, a black-box adversarial framework that probes model decision boundaries around \emph{similar} individuals. The key idea is to disentangle semantic content from surface form and test whether robustness and individual fairness are brittle under meaning-preserving lexical variation.
\textsc{RIFair} instantiates this idea via a two-stage pipeline: it first identifies influential tokens and then searches for admissible substitutions constrained by \emph{bidirectional entailment}~\cite{Truth-Conditional_Semantics}, thereby ensuring that perturbed inputs preserve the original truth conditions.
A core design choice in \textsc{RIFair} is \emph{decoupled substitution}: it applies distinct yet semantically equivalent substitutions to each member of a similar pair (e.g., mapping \texttt{talk} $\to$ \texttt{discuss} for one input and $\to$ \texttt{chat} for the other; Figure~\ref{fig:UnRobust_or_Unfair}). This decoupling isolates within-pair surface-form differences and, in turn, exposes robustness and fairness failures attributable to lexical variation rather than semantic drift.

Figure~\ref{fig:UnRobust_or_Unfair} visualizes  RB/UF/UB modes exposed by \textsc{RIFair}. The left column reports clean similar pairs that differ only in gendered tokens (e.g., his$\leftrightarrow$her, actor$\leftrightarrow$actress), whereas the right column presents their semantic-preserving adversarial counterparts, whose surface-form variations trigger distinct behaviors. Row~1 exemplifies \emph{RB} (robust yet biased): meaning-equivalent lexical substitutions (e.g., \textit{talk}$\to$\textit{discuss}/\textit{chat}) preserve correctness for one instance while inducing within-pair inconsistency, which evades robustness-only checks. Row~2 illustrates \emph{UF}: both predictions become incorrect relative to the ground truth yet remain consistent across the pair, a failure pattern that evades fairness-only checks. Row~3 corresponds to \emph{UB}, the case where predictions are simultaneously unrobust and biased, which many prior evaluations can detect. \textsc{RIF} is designed to capture all three regimes within a unified framework.

\paragraph{Empirical summary.}
Across diverse model architectures and real-world textual datasets, our experiments show that models often exhibit compound failures under perturbations, compromising robustness, individual fairness, or both simultaneously. As a result, robustness-only and fairness-only metrics frequently miss important regimes and fail to detect the multi-dimensional adversarial pairs generated by \textsc{RIFair}. Moreover, the bidirectional-entailment constraints preserve original truth conditions, yielding adversarial instances that remain imperceptible to humans while inducing model failures. By quantifying resilience over the joint robustness–fairness threat landscape, \textsc{RIF} captures these multi-dimensional vulnerabilities. These findings suggest that robustness and fairness should be evaluated jointly for trustworthy decision-making in high-stakes settings.

\paragraph{Contributions.}
This work makes four main contributions:
\textbf{(1) Unified criterion.} We formalize \textsc{RIF}, a unified notion that jointly characterizes robustness and individual fairness under semantic-preserving perturbations.
\textbf{(2) Semantic rigor.} We ground semantic-preserving perturbations in formal semantics via \emph{bidirectional entailment}, ensuring that admissible substitutions preserve the original truth conditions.
\textbf{(3) Black-box adversarial framework.} We propose \textsc{RIFair}, a black-box framework with a \emph{decoupled substitution} strategy that exposes robustness and fairness violations induced by superficial lexical variation.
\textbf{(4) Empirical evidence.} We provide extensive empirical evidence that robustness-only and fairness-only metrics miss key coupled failure modes, motivating joint evaluation under multi-dimensional threats.

\section{Related Work}

\textbf{Perturbations and robustness.}
Adversarial perturbations can cause DNNs to produce incorrect predictions while remaining nearly imperceptible to humans~\cite{FGSM, ma2026safetyscalecomprehensivesurvey,zhu2024attexplore,10.1007/978-3-032-06109-6_10}. This observation has motivated extensive research on \emph{adversarial robustness} \cite{abushaqra2024seqlink,chen2025ads, 9978964, 10.1145/3748503}, which characterizes a model's ability to maintain predictive accuracy under worst-case perturbations~\cite{R3}.
Robustness is commonly evaluated using \emph{white-box} attacks (e.g., FGSM~\cite{FGSM}, PGD, and stronger automated variants such as APGD~\cite{APGD} and ACG~\cite{ACG}) that exploit gradient information to search for worst-case perturbations. In practical deployments, adversaries may have only query access to a model, motivating \emph{black-box} attacks that estimate gradients from queries or perform gradient-free search (e.g., TextFooler~\cite{textfooler}).

Most robustness evaluations emphasize instance-level accuracy and do not assess whether perturbations disrupt \emph{consistency across similar individuals}. Consequently, a model may appear robust under standard evaluations while still exhibiting inequitable or discriminatory responses under perturbations.

\textbf{Perturbations and fairness.}
Adversarial perturbations can affect fairness at both the group and individual levels~\cite{RobustnessBias1,trustworthiness_1,ADF1, 10.1007/s10618-025-01112-8, DBLP:conf/cikm/FangZM25, 11043132, 10.1145/3630106.3658989}. At the group level, prior work shows that perturbations can exacerbate disparities by making some subpopulations systematically more (or less) robust than others---a phenomenon termed \emph{robustness bias}~\cite{RobustnessBias}.
At the individual level, perturbations may induce disparate treatment among similar individuals. Many approaches assume a \emph{white-box} setting and maximize prediction discrepancies under a user-specified similarity relation (e.g., ADF~\cite{ADF}, EIDIG~\cite{EIDIG}, and DICE~\cite{DICE}). By contrast, \emph{black-box} fairness testing targets API-only deployments and relies on lightweight gradient estimation from queries to guide attribute perturbations and discover individual-fairness violations (e.g., MAFT~\cite{MAFT}).

Many fairness evaluations are designed to \emph{surface discrimination} by amplifying prediction differences between similar inputs, without explicitly enforcing meaning preservation. Consequently, generated instances may drift semantically and even change the ground truth; moreover, because ground-truth labels are typically not incorporated during generation, it becomes difficult to jointly assess predictive correctness (robustness) and fairness. Finally, many methods are tailored to tabular settings (e.g., flipping gender or race), which is difficult to justify for natural-language inputs where meaning-preserving counterparts are substantially harder to define.

\emph{Limitations of treating robustness and fairness separately.}
Robustness and fairness are often evaluated in isolation, even though perturbations can influence both dimensions simultaneously. Consequently, a model may be robust yet unfair, or fair yet unrobust. We address this gap by introducing \textsc{RIF}, which jointly evaluates robustness and individual fairness under \emph{semantic-preserving} perturbations, and by proposing \textsc{RIFair} to surface unrobust and/or unfair similar-instance pairs.

\section{Robust Individual Fairness}

We propose \textsc{RIF} as a joint trustworthiness principle: systems should not produce materially different outcomes for inputs that are equivalent in task-relevant information, particularly when differences are driven by sensitive attributes or superficial surface-form variation rather than substantive changes.

\emph{Setup:}
Let $\mathcal{V}$ and $\mathcal{Y}$ denote the input and label spaces. We consider a labeled dataset $D \subseteq \mathcal{V} \times \mathcal{Y}$, where each instance $v \in \mathcal{V}$ is a token sequence $v=(x_1,\dots,x_n)$ with label $y \in \mathcal{Y}$. Let $\mathcal{A}$ be a set of sensitive-attribute tokens (e.g., race or gender identifiers). We write $x_i \in \mathcal{A}$ when token $x_i$ is sensitive. A classifier $f: \mathcal{V} \to \mathcal{Y}$ outputs a prediction $\hat{y}=f(v)$.

\subsection{Foundation: Semantic Equivalence}
\label{sec:preliminaries}

Robustness and fairness evaluations frequently define perturbations and similar individuals using geometric neighborhoods (e.g., $L_p$ balls, cosine similarity). In NLP, however, geometric proximity (e.g., cosine similarity) is an unreliable proxy for meaning preservation: it admits merely topically related substitutions~\cite{cosine_metric} that may alter truth conditions. Since our goal is to study robustness--fairness interactions under \emph{meaning-preserving} edits, we first formalize semantic equivalence and then specify its operational approximation.

\emph{Ideal notion (truth-conditional semantics).}
Following formal semantics~\cite{Truth-Conditional_Semantics}, we treat the meaning of an expression $v$ as its truth set over possible worlds $\mathcal{W}$, denoted $\llbracket v \rrbracket \subseteq \mathcal{W}$. Two expressions $v$ and $v'$ are semantically equivalent if and only if they have identical truth conditions:
\begin{equation}
\label{eq:semantic_equivalence}
\text{Sem}(v) \equiv \text{Sem}(v') \iff \llbracket v \rrbracket = \llbracket v' \rrbracket.
\end{equation}

\emph{Operational notion (entailment for semantic identity).}
Since directly reasoning over (potentially infinite) possible worlds is infeasible, we operationalize semantic identity via entailment patterns~\cite{Truth-Conditional_Semantics}.
In particular, we adopt \emph{bidirectional entailment} as the criterion for semantic equivalence: an expression $v'$ represents the same underlying semantic entity as $v$ if and only if $v$ entails $v'$ and $v'$ entails $v$.
Formally,
\begin{equation*}
\label{eq:semantic_identity_entailment}
\llbracket v \rrbracket = \llbracket v' \rrbracket \iff (v \models v') \wedge (v' \models v).
\end{equation*}
We instantiate $\models$ using a strong natural language inference (NLI) model and treat the resulting bidirectional entailment score as a high-precision admissibility filter.

\emph{Assumptions:}
\begin{itemize}
    \item \textbf{A1 (Label invariance).} For the datasets in this work, admissible meaning-preserving edits do not change the ground-truth label $y$.
    \item \textbf{A2 (Semantic filter).} A candidate edit is admissible if it passes a bidirectional entailment check under a strong natural language inference (NLI) model~\cite{NFI_model}.
\end{itemize}

For a pair $(v,v')$ we define the semantic-equivalence score
\begin{equation*}
\label{eq:seq_def}
S_{\mathrm{eq}}(v,v') \;=\; P(v \models v')\, P(v' \models v),
\end{equation*}
where $P(\cdot)$ is the NLI entailment probability. We accept $v'$ as meaning-preserving for $v$ when $S_{\mathrm{eq}}(v,v') \ge \tau_{\mathrm{sem}}$.

\emph{From semantic equivalence to similar individuals and admissible perturbations.}
We use the semantic filter to define  (i) the semantic set of similar individuals and (ii) the space of admissible textual perturbations. Let $M(v,\mathcal{A})$ be a masking operator that obscures tokens in $\mathcal{A}$. 
 The semantic set of $v$ is
\begin{equation*}
\label{eq:similarity_set}
I(v) \;=\; \bigl\{v' \in \mathcal{V} \;\big|\; S_{\mathrm{eq}}\bigl(M(v,\mathcal{A}),\, M(v',\mathcal{A})\bigr) \ge \tau_{\mathrm{sem}}\bigr\}.
\end{equation*}

In addition, we call a token-level substitution $x_i \rightarrow x_i'$ \emph{semantically imperceptible} if $S_{\mathrm{eq}}(x_i,x_i') \ge \tau_{\mathrm{sem}}$; such substitutions induce an admissible perturbation space $\Delta_{\text{adm}}$. This construction aims to isolate failures due to model sensitivity under meaning preservation, rather than label-changing semantic drift.

\subsection{RIF Definition}

Under A1--A2, we define \textsc{RIF} as a universal robustness property over the semantic set $I(v)$.

\begin{definition}[Robust Individual Fairness]
\label{RIF-definition}
A classifier $f$ satisfies \emph{$\epsilon$-Robust Individual Fairness} (\emph{$\epsilon$-RIF}) on an instance $v$ with label $y$ if, for every similar counterpart $v' \in I(v)$ and every admissible perturbation $\delta \in \Delta_{\text{adm}}$, the prediction error remains bounded by $\epsilon$:
\begin{equation}
\label{eqn:RIF_unified}
\forall v' \in I(v),\; \forall \delta \in \Delta_{\text{adm}}:\quad D\bigl(y, f(v' \oplus \delta)\bigr) \le \epsilon,
\end{equation}
where $D(\cdot,\cdot)$ is a loss/distance metric (e.g., cross-entropy), $\oplus$ denotes applying a textual perturbation, and $\epsilon \ge 0$ is a tolerance parameter.
\end{definition}

Intuitively, Eq.~\ref{eqn:RIF_unified} requires correctness to be stable not only under admissible perturbations of $v$, but uniformly over all counterfactual counterparts in $I(v)$. As a result, \textsc{RIF} rules out failures that are missed when robustness and individual fairness  are assessed separately.

\subsection{ The Robustness - Fairness Relationship}
\label{sec:rif_analysis}


\begin{figure*}[t]
    \centering
    \includegraphics[width=1.\linewidth]{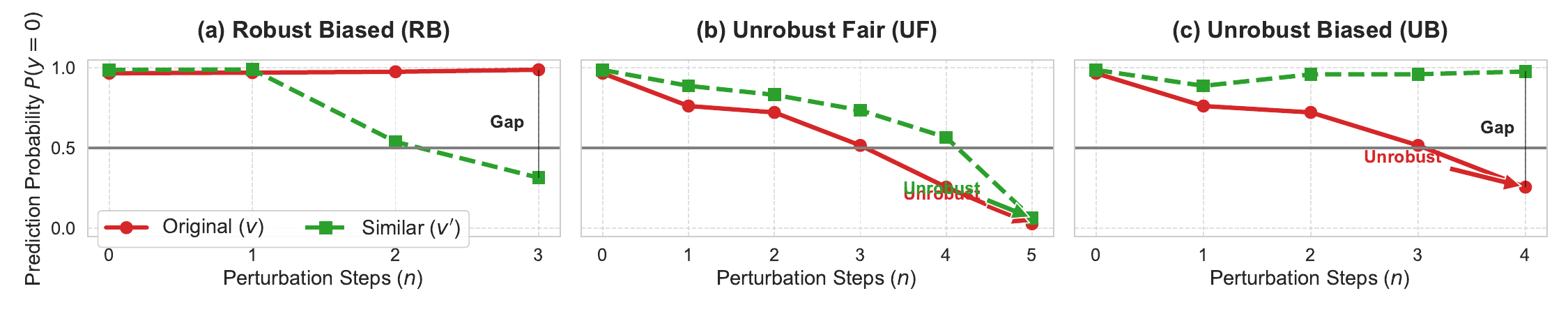}
    \caption{Prediction-change trajectories for similar instances under targeted RB/UF/UB objectives.}
    \label{fig:three_attack_change}
\end{figure*}

Robustness (\emph{R}) and individual fairness (\emph{F}) capture distinct behaviors and can vary independently. Figure~\ref{fig:R F Relationship} summarizes the four qualitatively different quadrants; notably, robustness-only evaluations may accept \emph{Robust Biased} (RB) behavior, while fairness-only evaluations may accept \emph{Unrobust Fair} (UF) behavior.

The following theorem makes the link between \textsc{RIF} and the isolated notions explicit.

\begin{theorem}[Interplay of RIF, Robustness, and Fairness]
\label{thm:rif_characterization}
Let $f: \mathcal{V} \to \mathcal{Y}$ be a classifier and $D(\cdot,\cdot)$ satisfy the triangle inequality.

\noindent \textbf{1. Necessity (RIF $\Rightarrow$ R and F).}
If $f$ satisfies $\epsilon$-RIF on $v$, then for any admissible perturbation $\delta$ it holds that
\begin{align*}
\text{(i) Robustness:}\quad & D\bigl(y, f(v \oplus \delta)\bigr) \le \epsilon, \\
\text{(ii) Individual fairness:}\quad & D\bigl(f(v \oplus \delta), f(v' \oplus \delta)\bigr) \le 2\epsilon,\quad \forall v' \in I(v).
\end{align*}

\noindent \textbf{2. Sufficiency with relaxation (R and F $\Rightarrow$ relaxed RIF).}
If $f$ satisfies robustness $D(y, f(v\oplus\delta)) \le \epsilon$ and Lipschitz-style fairness $D(f(v\oplus\delta), f(v'\oplus\delta)) \le K\, d(v\oplus\delta, v'\oplus\delta)$, then
\begin{equation*}
D\bigl(y, f(v'\oplus\delta)\bigr) \le \epsilon + K\, d(v\oplus\delta, v'\oplus\delta).
\end{equation*}
\text{Proof details are deferred to Appendix~\ref{ProofSection3}.}
\end{theorem}

\noindent\textbf{Discussion.}
Theorem~\ref{thm:rif_characterization} clarifies that (i) universal robustness over $I(v)$ implies a bounded fairness guarantee (with a $2\epsilon$ factor), and (ii) checking robustness and fairness in isolation implies only a relaxed version of \textsc{RIF}. This explains why the RB and UF quadrants can evade conventional metrics.

\section{Adversarial Similar Instances Generation}
\label{sec:rifair_generation}

We propose \textsc{RIFair}, a black-box adversarial framework for \emph{constructing semantically preserved} similar-instance pairs that nevertheless violate \textsc{RIF}. Given a semantically matched pair $(v, v')$ (Section~\ref{sec:preliminaries}), \textsc{RIFair} searches for \emph{decoupled} admissible perturbations $(\delta, \delta') \in \Delta_{\text{adm}}\times\Delta_{\text{adm}}$ that induce (i) robustness failures, (ii) individual-fairness violations, or (iii) both.

\subsection{Unified Objective}
\label{sec:unified_objective}

We cast adversarial generation as a joint optimization problem over robustness and fairness. Specifically, we minimize
\begin{equation}
\label{eq:unified_objective_simplified}
\begin{aligned}
\min_{\delta, \delta' \in \Delta_{\text{adm}}} \; \mathcal{J}(\delta,\delta')
= \;&
\underbrace{\alpha_v \, \mathcal{L}\bigl(f(v \oplus \delta), y\bigr)
+ \alpha_{v'} \, \mathcal{L}\bigl(f(v' \oplus \delta'), y\bigr)}_{\text{Robustness term}} \\
&\quad +
\underbrace{\beta \, \mathcal{F}\bigl(f(v \oplus \delta), f(v' \oplus \delta')\bigr)}_{\text{Fairness term}} ,
\end{aligned}
\end{equation}
where $\mathcal{L}$ is the cross-entropy loss with respect to the ground truth $y$, and $\mathcal{F}$ is a discrepancy functional between predictions (e.g., $\ell_2$ distance between probability vectors). The sign parameters $\alpha_v, \alpha_{v'}, \beta \in \{+1,-1\}$ control whether each term is minimized or maximized.
By varying $(\alpha_v,\alpha_{v'},\beta)$, we can target the RB/UF/UB quadrants as shown in Table~\ref{tab:optimization_weights} and Figure \ref{fig:three_attack_change}.


\begin{table}[h]
\centering
\caption{Hyperparameters for RIF violations.}
\label{tab:optimization_weights}
\begin{threeparttable}
\resizebox{\columnwidth}{!}{%
\begin{tabular}{l|cc|c}
\toprule
\textbf{Target Violation} & $\alpha_v$ (Rob $v$) & $\alpha_{v'}$ (Rob $v'$) & $\beta$ (Fairness) \\ \midrule
\textbf{Robust Biased (RB)} & $+1$ (Min Loss) & $-1$ (Max Loss) & $-1$ (Max Gap) \\
\textbf{Unrobust Biased (UB)} & $-1$ (Max Loss) & $+1$ (Min Loss) & $-1$ (Max Gap) \\
\textbf{Unrobust Fair (UF)} & $-1$ (Max Loss) & $-1$ (Max Loss) & $+1$ (Min Gap) \\ \bottomrule
\end{tabular}%
}
\end{threeparttable}
\end{table}

\subsection{Perturbation-Impact Diagnostics}
\label{sec:perturbation_impact}

To interpret the dynamics induced by Eq.~\ref{eq:unified_objective_simplified}, we introduce two diagnostics that summarize the \emph{realized} (finite-step) effect of a discrete textual edit.
Let $p(v) \in \mathbb{R}^{|\mathcal{Y}|}$ denote the predictive distribution returned by $f$, and define $\Delta p \triangleq p(v \oplus \delta) - p(v)$. We define:
\begin{equation*}
\label{eq:pii_pid}
\mathrm{PII}(v,\delta) \;=\; \frac{\lVert \Delta p \rVert_2}{\lVert \delta \rVert},
\qquad
\mathrm{PID}(v,\delta) \;=\; \frac{\Delta p}{\lVert \Delta p \rVert_2} \quad (\lVert \Delta p \rVert_2 > 0).
\end{equation*}
where $\lVert \delta \rVert$ denotes perturbation magnitude under an edit-distance proxy (e.g., the number of substitutions). 
Perturbation Impact Index ($\mathrm{PII}$) measures empirical sensitivity under a concrete edit, while Perturbation Impact Direction  ($\mathrm{PID}$) specifies the direction of the induced movement in prediction space.

Accordingly, each perturbation step admits the decomposition
\begin{equation*}
\label{eq:single_decomposition}
\Delta p \;=\; \mathrm{PII}(v,\delta)\, \lVert \delta \rVert\, \mathrm{PID}(v,\delta).
\end{equation*}

For a sequence of edits $\{\delta_i\}$ applied to $v$, we obtain the cumulative expansion
\begin{equation}
\label{eq:cumulative_impact}
p\Bigl(v \oplus \sum_i \delta_i\Bigr) = p(v) + \sum_i \mathrm{PII}_i^{v}\, \lVert\delta_i\rVert\, \mathrm{PID}_i^{v}.
\end{equation}
Comparing the resulting $\mathrm{PII}$ and $\mathrm{PID}$ trajectories between $v$ and $v'$ reveals whether an attack succeeds primarily via \emph{heterogeneous sensitivity} (magnitude mismatch) or via \emph{directional misalignment}.

\emph{Role of $\alpha$ and $\beta$.}
Substituting Eq.~\ref{eq:cumulative_impact} into Eq.~\ref{eq:unified_objective_simplified} provides an intuitive interpretation of the objective weights.
\begin{itemize}
    \item \textbf{Sensitivity control ($\alpha$).} The robustness weights $\alpha_v$ and $\alpha_{v'}$ encourage ($\alpha=-1$) or suppress ($\alpha=+1$) high-impact edits, modulating the magnitude of $\mathrm{PII}$ along each trajectory.
    \item \textbf{Alignment control ($\beta$).} The fairness weight $\beta$ promotes alignment ($\beta=+1$) or divergence ($\beta=-1$) between the directions of prediction change, i.e., between $\mathrm{PID}^v$ and $\mathrm{PID}^{v'}$.
\end{itemize}

\emph{Connecting objectives to quadrants.}
The three hyperparameter settings in Table~\ref{tab:optimization_weights} correspond to distinct trade-offs between robustness and fairness in Eq.~\ref{eq:unified_objective_simplified} (Figure~\ref{fig:three_attack_change}).
\textbf{Robust Biased (RB).} $(\alpha_v=+1,\alpha_{v'}=-1,\beta=-1)$ stabilizes $v$ while destabilizing $v'$ and encourages prediction divergence; violations are primarily driven by \emph{heterogeneous sensitivity}.
\textbf{Unrobust Fair (UF).} $(\alpha_v=-1,\alpha_{v'}=-1,\beta=+1)$ destabilizes both instances but enforces alignment, so the pair tends to fail \emph{together}: large error with a small prediction gap.
\textbf{Unrobust Biased (UB).} $(\alpha_v=-1,\alpha_{v'}=+1,\beta=-1)$ destabilizes $v$ while stabilizing $v'$ and encourages divergence, yielding simultaneous robustness failure and individual-fairness violation.

\subsection{\textsc{RIFair} Algorithm}
\label{sec:rifair_algorithm}

Algorithm~\ref{alg:rifair} instantiates Eq.~\ref{eq:unified_objective_simplified} in a query-only setting. Our central design choice is \emph{decoupling}: unlike coupled attacks that enforce identical substitutions on $v$ and $v'$, \textsc{RIFair} explores the admissible space independently for each instance. This is critical for surfacing RB, UB, and UF behaviors that arise from asymmetric surface forms.

\begin{algorithm}[h]
\caption{{RIFair}: Black-Box Decoupled Adversarial Generation}
\label{alg:rifair}
\begin{algorithmic}[1]
\REQUIRE Similar pair $(v, v')$; label $y$; model $f$; weights $\alpha_v, \alpha_{v'}, \beta$
\ENSURE Adversarial pair $(v_{adv}, v'_{adv})$ satisfying the target violation

\STATE \textbf{Initialize:} $v_{adv} \leftarrow v$, $v'_{adv} \leftarrow v'$, $J^* \leftarrow \mathcal{J}(v, v')$

\STATE \COMMENT{\textbf{Stage 1: query-based importance ranking}}
\FOR{each token $x_k$ in $v$}
    \STATE $I[x_k] \leftarrow |f(v) - f(v_{\setminus x_k})|$ \COMMENT{Occlusion saliency}
\ENDFOR
\STATE $R_{ranked} \leftarrow$ tokens sorted by $I[x_k]$ in descending order

\STATE \COMMENT{\textbf{Stage 2: iterative decoupled substitution}}
\FOR{each token $x_i$ in $R_{ranked}$} 
\STATE Generate semantic-preserving candidates $S_v, S_v'$ from admissible perturbations $\Delta_{\mathrm{adm}}^{x_i}$
    \STATE $S_v \leftarrow \{v_{{x_i} \oplus \delta} \mid \delta \in \Delta_{adm}^{x_i}\}$, \; $S_{v'} \leftarrow \{v'_{{x_i} \oplus \delta'} \mid \delta' \in \Delta_{adm}^{x_i}\}$

    \STATE Query model: $\mathbf{P}_v \gets f(S_v)$, \; $\mathbf{P}_{v'} \gets f(S_{v'})$

    \STATE $\mathcal{I},\mathcal{K} \gets \{\arg_{\min}(\mathbf{P}_v), \arg_{\max}(\mathbf{P}_v)\}, \{\arg_{\min}(\mathbf{P}_{v'}), \arg_{\max}(\mathbf{P}_{v'})\}$

    \STATE $(i^*,j^*) \leftarrow \arg_{\min\limits_{(i,j) \in \mathcal{I} \times \mathcal{K}}}
    \Bigl[\beta D(\mathbf{P}_v[i], \mathbf{P}_{v'}[j]) + \alpha_v \mathcal{L}(\mathbf{P}_v[i], y) + \alpha_{v'} \mathcal{L}(\mathbf{P}_{v'}[j], y)\Bigr]$

    \STATE Compute $\mathcal{J}_{x_{i}}$ using $(i^*, j^*)$
    \IF{$\mathcal{J}_{x_{i}} < J^*$}
        \STATE $J^* \leftarrow \mathcal{J}_{x_{i}}$; $v_{adv} \leftarrow S_v[i^*]$; $v'_{adv} \leftarrow S_{v'}[j^*]$
    \ENDIF

    \IF{Success($f(v_{adv}), f(v'_{adv}), y$)}
        \RETURN $(v_{adv}, v'_{adv})$
    \ENDIF
\ENDFOR

\RETURN $(v_{adv}, v'_{adv})$
\end{algorithmic}
\end{algorithm}

\textbf{Implementation details:} \emph{Query efficiency.} Stage~1 (lines~4--6) ranks tokens by occlusion saliency using model queries only. \emph{Efficient pair perturbation.} Stage~2 (lines~9--15) constructs semantic-preserving candidate sets $S_v$ and $S_{v'}$ from admissible perturbations $\Delta_{\mathrm{adm}}^{x_i}$ and minimizes $\mathcal{J}_{x_i}$ by evaluating only extreme candidates (min/max confidence) and their Cartesian product. \emph{Early stopping.} (lines~ 18-20) stops attack once the target-quadrant criterion is met.


\begin{table*}[h!]
\centering
\caption{Model Performance Comparison across RIF, Robustness (TextFooler), and Fairness (MAFT \& ADF) metrics.}
\label{tab:model_comparison_full}
\begin{tabular}{c|cc|cc|cc|cc}
\hline
\multirow{3}{*}{{Model}} & \multicolumn{2}{c|}{{RIF}} & \multicolumn{2}{c|}{{Robustness}} & \multicolumn{4}{c}{{Fairness}} \\
\cmidrule(lr){2-3} \cmidrule(lr){4-5} \cmidrule(lr){6-9}
 & \multicolumn{2}{c|}{\textit{RIFair}} & \multicolumn{2}{c|}{\textit{TextFooler}} & \multicolumn{2}{c|}{\textit{MAFT}} & \multicolumn{2}{c}{\textit{ADF}} \\
\cmidrule(lr){2-3} \cmidrule(lr){4-5} \cmidrule(lr){6-7} \cmidrule(lr){8-9}
 & {$RIF$} & {$N_{attack}$} & $R$ & $N_{attack}$ & $F$ & $N_{attack}$ & $F$ & $ N_{attack}$ \\
\hline
DeBERTa-v3 & \textbf{0.605} & 16.71 & 0.704 & 16.97 & 0.972 & 18.70 & 0.972 & 18.69 \\
BERT & 0.592 & 16.59 & 0.713 & 17.20 & \textbf{0.977} & 18.68 & \textbf{0.977} & 18.66 \\
RoBERTa & 0.580 & 16.62 & 0.700 & \textbf{16.90} & 0.976 & 18.69 & 0.974 & \textbf{18.62} \\
DistilBERT & 0.574 & \textbf{16.56} & \textbf{0.717} & {17.10} & 0.971 & \textbf{18.64} & 0.973 & 18.63 \\
\hline
\textit{Average} & \textbf{0.588} & \textbf{16.62} & 0.709 & 17.04 & 0.974 & 18.68 & 0.974 & 18.65 \\ \hline
\end{tabular}%
\end{table*}

\subsection{Implementing Semantic Equivalence}
\label{sec:semantic_equivalence}
We implement the similarity set $I(v)$ and the admissible perturbation set $\Delta_{\text{adm}}$ using two tailored generation strategies.

\emph{1. Constructing the similarity set $I(v)$ via masking.}
Recall:
\begin{equation*}
    I(v) = \{ v' \in \mathcal{V} \mid S_{\mathrm{eq}}(M(v, \mathcal{A}), M(v', \mathcal{A})) \ge \tau_{\mathrm{sem}} \}.
\end{equation*}
We construct candidates $v'$ by enforcing \emph{ceteris paribus} changes on sensitive attributes only. Specifically, $M(v,\mathcal{A})$ identifies sensitive tokens from a predefined lexicon $\mathcal{A}$ (e.g., gendered pronouns or racial identifiers) and replaces them with typed placeholders. We then generate counterparts by filling each sensitive slot with alternative values from the corresponding attribute domain.
This masking-based construction guarantees that the non-sensitive content of $v$ and $v'$ is identical. Consequently, their masked forms match exactly, yielding a maximal semantic-equivalence score $S_{\mathrm{eq}}(M(v,\mathcal{A}), M(v',\mathcal{A}))$ and satisfying the threshold by design.

For example, if $v$ contains `\emph{he} is a nurse', then $M$ masks `he', and we construct $v'$ by substituting alternatives such as `she' or `they' at the \emph{same position}. This keeps $v$ and $v'$ structurally identical, ensuring that the sensitive attribute is the only varying factor. Pseudocode and additional examples are provided in Appendix~\ref{sec:mask-function}.

\begin{figure}[h]
    \centering
    \includegraphics[width=1.\linewidth]{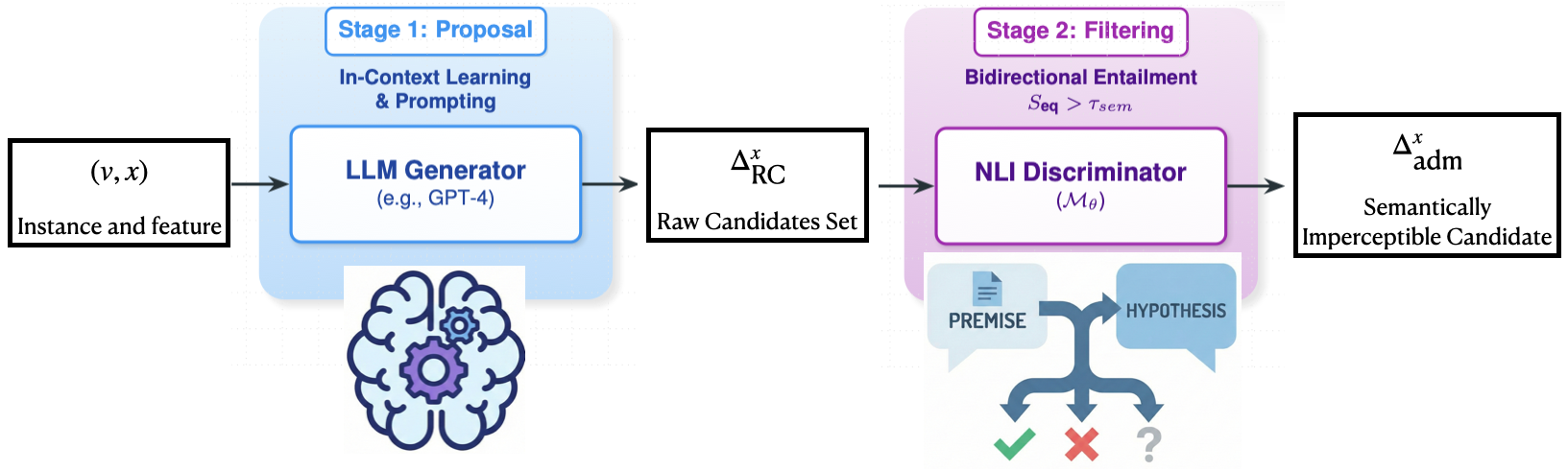}
    \caption{Generate--Filter pipeline }
    \label{fig:pipeline}
\end{figure}

\emph{2. Constructing perturbations $\Delta_{\text{adm}}$ via Generate--Filter.}
Adversarial perturbations require diverse surface-form realizations while preserving meaning. We instantiate $\Delta_{\text{adm}}$ via a two-stage \emph{Generate--Filter} pipeline (Figure~\ref{fig:pipeline}) that prioritizes semantic preservation.

\emph{\textbf{Stage 1 (LLM generator).}} For a target token or phrase $x_i$ in $v$, we use a large language model to propose a candidate set of paraphrases/synonyms.

\emph{\textbf{Stage 2 (NLI discriminator).}}  The Candidates is filtered by a NLI model \cite{NFI_model} and keep only those satisfying bidirectional entailment, i.e., $S_{\mathrm{eq}}(x_i,x_i') \ge \tau_{\mathrm{sem}}$.

Concretely, we compute token frequencies in the attack corpus and select the top 1{,}000 most frequent tokens as substitution targets. For each selected token, we query the LLM to propose 10--20 candidate synonyms. The generation prompt and filtering procedure are detailed in Appendix~\ref{sec:semantic-admissible-perturbation}.

\section{Experimental Evaluation}

\paragraph{Implementation.}
We implement \textsc{RIFair} in Python and run all experiments on NVIDIA Tesla 2 $\times$ T4 GPUs.
For the \emph{Generate--Filter} pipeline (Fig.~\ref{fig:pipeline}), we use Qwen2.5-7B \cite{qwen2} as the candidate generator and DeBERTa-v3-Large \cite{NFI_model} as the NLI filter, with setting $\tau_{\text{sem}}=0.8$.

\emph{Route map.}
We first describe datasets, models, and evaluation metrics; we then (i) quantify robustness--fairness misalignment under multi-perturbation threats, (ii) analyze metric sensitivity across RB/UF/UB scenarios, (iii) visualize trajectory-level dynamics on similar pairs, and (iv) report defenses and ablations.

\subsection{Experimental Setup}

\noindent\textbf{\emph{Datasets.}}
We evaluate on three socially sensitive NLP benchmarks---Bias in Bios~\cite{Bias_in_Bios}, Jigsaw~\cite{jigsaw}, and SST-2~\cite{SST-2}---and one tabular decision-making benchmark, \textit{Census Income (Adult)}~\cite{adult}.
Sensitive tokens include gender, sexuality, religion, race, and disability. For each dataset, we identify sensitive attributes and construct semantic counterparts using the masking procedure in Section~\ref{sec:semantic_equivalence}.
To enable textual fairness evaluation for tabular records, we use Qwen2.5~\cite{qwen2} to verbalize each Adult record into a natural-language description.

\noindent\textbf{\emph{Models.}}
We study four transformer encoders: BERT~\cite{bert}, RoBERTa~\cite{RoBERTa}, DistilBERT~\cite{DistilBERT}, and DeBERTa-v3~\cite{NFI_model}.

\noindent\textbf{\emph{Metrics.}}
We report three perturbation-based resistance rates: robustness $R$, individual fairness $F$, and our joint criterion $RIF$.
For explainability, we additionally report the failure rates corresponding to the three non-RF quadrants in Figure~\ref{fig:R F Relationship}: Robust Biased ($R_{RB}$), Unrobust Fair ($R_{UF}$), and Unrobust Biased ($R_{UB}$).

\subsection{Multiple-Perturbation Threats}
\label{sec:method_analysis}

\begin{table*}[h]
\centering
\caption{{Comprehensive Analysis \& Metric Sensitivity}}
\label{tab:combined_results}
\begin{tabular}{c|ccc|ccc|ccc}
\hline
\multirow{2}{*}{{Model}} & \multicolumn{3}{c|}{{Target: RB}} & \multicolumn{3}{c|}{{Target: UF}} & \multicolumn{3}{c}{{Target: UB}} \\ 
\cmidrule(lr){2-4} \cmidrule(lr){5-7} \cmidrule(lr){8-10}
 & $R$ & $F$ & $RIF$ & $R$ & $F$ & $RIF$ & $R$ & $F$ & $RIF$ \\ \hline
 {BERT} & 0.942 & 0.824 & 0.766 & 0.700 & 0.958 & 0.698 & 0.710 & 0.799 & 0.710 \\
{RoBERTa} & 0.932 & 0.814 & 0.746 & 0.713 & 0.951 & 0.696 & 0.733 & 0.801 & 0.733 \\
{DistilBERT} & 0.938 & 0.800 & 0.738 & 0.707 & 0.953 & 0.690 & 0.720 & 0.787 & 0.720 \\
{DeBERTa-v3} & 0.914 & 0.836 & 0.755 & 0.730 & 0.895 & 0.686 & 0.755 & 0.832 & 0.752 \\ \hline
\textbf{Average} & 0.932 & 0.819 & \textbf{0.751} & 0.713 & 0.939 & \textbf{0.693} & 0.730 & 0.805 & \textbf{0.729} \\ \hline
\end{tabular}%
\end{table*}
We evaluate (i) robustness under TextFooler (metric $R$), (ii) individual fairness under MAFT/ADF (metric $F$), and (iii) the joint criterion under \textsc{RIFair} (metric $RIF$).
Since ADF is originally gradient-based and designed for tabular inputs, we adapt it to the black-box NLP setting by replacing gradients with query-based token-importance estimates (TextFooler-style occlusion) to guide perturbation generation.
To control for candidate quality and search-space size, all methods are constrained to perturb only the top-20 most important tokens per input and to use the same semantic-admissibility threshold for substitution candidates.
We report the resistance rate (higher is better) and the average attack cost $N_{attack}$ (lower is better), defined as the number of model queries required to induce a violation under a fixed budget.

Table~\ref{tab:model_comparison_full} demonstrates a robustness--fairness misalignment under token perturbations.
Fairness resistance is uniformly high across models, whereas robustness resistance is consistently lower; the joint resistance $RIF$ is the most stringent criterion (lowest values), reflecting simultaneous vulnerability to robustness failures and/or fairness violations.
Crucially, model rankings depend on the evaluation axis: DistilBERT attains the highest robustness score ($R$), BERT attains the highest fairness score ($F$), yet DeBERTa-v3 achieves the strongest joint resistance ($RIF$).

Regarding efficiency, robustness attacks are typically more query-efficient than fairness attacks (lower $N_{attack}$) because $R$ optimizes a single input, whereas $F$ optimizes over a semantically matched pair.
In contrast, \textsc{RIFair} often requires fewer queries because it succeeds whenever \emph{either} robustness fails or fairness is violated, i.e., under a strictly broader set of failure conditions.

\subsection{Metric Sensitivity}
\label{sec:metric_analysis}

Table~\ref{tab:combined_results} reports resistance under three targeted scenarios (RB, UF, UB) and demonstrates a scenario-dependent misalignment between $R$ and $F$.
We use \emph{metric sensitivity} to denote whether a metric changes substantially when a given failure mode is present.

\textbf{Single-axis limitations.}
Robustness $R$ is primarily sensitive to instability (UF/UB) and can be insensitive in RB when each individual prediction remains stable but a discriminatory gap emerges.
Conversely, fairness $F$ is primarily sensitive to disparity (RB/UB) and can be insensitive in UF when both individuals drift together (remaining outcome-consistent) despite becoming jointly incorrect.

\textbf{Joint sensitivity of \textsc{RIF}.}
By jointly penalizing instability and disparity, $RIF$ remains consistently sensitive across RB/UF/UB.
This is reflected in the average row of Table~\ref{tab:combined_results}: under UF, $R$ drops sharply ($0.713$) while $F$ remains high ($0.939$); under RB, $R$ remains high ($0.932$) while $F$ degrades ($0.819$); and under UB both degrade.
Across all scenarios, $RIF$ provides a unified signal (lowest resistance in each scenario).

\subsection{Prediction Trajectories of Similar Pairs}

Figure \ref{fig:three_attack_change} traces prediction trajectories under RB/UF/UB objectives.
In RB (Panel~a), one instance remains stable while its counterpart undergoes a sharp confidence drop and crosses the decision boundary, creating an outcome gap.
In UB (Panel~c), this pattern is reversed.
In UF (Panel~b), both instances fail synchronously; although the prediction gap remains small, the behavior constitutes a robustness failure.
These dynamics illustrate why a joint criterion is needed: violations can manifest as disparity (RB/UB) or as joint instability (UF), which single-axis metrics cannot consistently detect.

\subsection{RIFair Perturbation Defense}

Table~\ref{tab:defense_results} reports adversarial retraining results on the Bias in Bios benchmark under the \textsc{RIF} threat model.
We use \textsc{RIFair} to generate RB/UF/UB adversarial instances from 1{,}000 \emph{training} examples (to avoid test leakage), then augment the training set, and retrain each backbone.
We then evaluate the retrained models on \textsc{RIFair}-generated test-time pairs.

\begin{table}[h]
\centering
\caption{\textbf{Defense Effectiveness.}}
\label{tab:defense_results}
\resizebox{1\linewidth}{!}{%
\begin{tabular}{l|ccc|ccc}
\toprule
\multirow{2}{*}{{Model}} & \multicolumn{3}{c|}{{Original Rate (\%)}} & \multicolumn{3}{c}{{Retrained Rate (Change)}} \\
\cmidrule(lr){2-4} \cmidrule(lr){5-7}
 & $R_{RB}$ & $R_{UF}$ & $R_{UB}$ & $R_{RB}$ & $R_{UF}$ & $R_{UB}$ \\ \midrule
BERT & 24.9 & 43.7 & 34.5 & \textbf{13.8} ($-$11.0) & \textbf{31.7} ($-$12.1) & \textbf{14.9} ($-$19.6) \\
RoBERTa & 26.1 & 42.3 & 29.5 & \textbf{12.7} ($-$13.4) & \textbf{28.8} ($-$13.6) & \textbf{15.2} ($-$14.3) \\
DistilBERT & 27.2 & 42.7 & 28.7 & \textbf{14.9} ($-$12.3) & \textbf{31.9} ($-$10.7) & \textbf{16.3} ($-$12.4) \\
DeBERTa-v3 & 25.2 & 42.9 & 33.2 & \textbf{12.4} ($-$12.8) & \textbf{29.9} ($-$13.1) & \textbf{15.4} ($-$17.9) \\ \midrule
\textbf{Average} & \textbf{25.9} & \textbf{42.9} & \textbf{31.5} & \textbf{13.5} ($-$12.4) & \textbf{30.6} ($-$12.4) & \textbf{15.5} ($-$16.1) \\ \bottomrule
\end{tabular}%
}
\end{table}

Adversarial retraining consistently reduces failure rates across all categories.
On average, $R_{RB}$ decreases from 25.9\% to 13.5\% ($\Delta=-12.4$), $R_{UF}$ from 42.9\% to 30.6\% ($\Delta=-12.4$), and $R_{UB}$ from 31.5\% to 15.5\% ($\Delta=-16.1$).
Overall, these gains indicate that re-training with \textsc{RIFair}-generated instances substantially improves robustness and individual fairness under the \textsc{RIF} threat model.
\subsection{Method Ablations Analysis}

A core requirement of \textsc{RIF} is that admissible perturbations $\Delta_{\text{adm}}$ are \emph{meaning-preserving}. Accordingly, we assess semantic preservation at two levels: (i) the \emph{substitution token candidates} produced by our \emph{Generate--Filter} pipeline and (ii) the \emph{adversarial instance pairs} constructed from these token candidates.

\begin{table*}[h]
\centering
\small
\caption{\textbf{Adversarial Pair Quality Comparison.}}
\label{tab:semantic_validity}
\setlength{\tabcolsep}{5pt}
\begin{tabular}{l|cccc|cccc}
\toprule
\multirow{2}{*}{\textbf{Model}} & \multicolumn{4}{c|}{{Ours: Semantic Substitution  (Entailment)}} & \multicolumn{4}{c}{{Baseline: Geometric Substitution  (Cosine)}} \\ 
\cmidrule(lr){2-5} \cmidrule(lr){6-9}
 & {RIF} & {Cosine} & {$S_{eq}$}  & $S_{judge}$  & {RIF} & {Cosine} & {$S_{eq}$}  & $S_{judge}$ \\ 
\midrule
BERT & \textbf{0.701} & 0.975 & 0.974 & \textbf{4.63} & 0.560 & 0.962 & 0.338 & 3.36 \\
DeBERTa-v3 & 0.688 & 0.962 & \textbf{0.981} & 4.59 & 0.471 & 0.949 & 0.379 & 3.45 \\
DistilBERT & 0.688 & 0.967 & 0.980 & 4.57 & 0.527 & 0.954 & 0.268 & 3.43 \\
RoBERTa & 0.700 & 0.962 & 0.976 & 4.60 & 0.492 & 0.955 & 0.333 & 3.43 \\ 
\midrule
\textbf{Average} & \textbf{0.694} & \textbf{0.967} & \textbf{0.978} & \textbf{4.60} & 0.513 & 0.955 & 0.330 & 3.42 \\ 
\bottomrule
\end{tabular}
\end{table*}

\subsubsection{Semantic Preservation of Substitution Candidates}
\label{sec:semantic_validity_ablation}
To obtain a more complete view of semantic preservation for substitution candidates, we compare three signals: geometric proximity (cosine similarity in embedding space), semantic equivalence via bidirectional entailment ($S_{eq}$), and a human-aligned LLM-as-a-judge score ($S_{judge}$).
We compute cosine similarity using \texttt{all-mpnet-base-v2}, a sentence-transformers model~\cite{cosine_simialrity_embeeding}, and compute $S_{eq}$ via bidirectional entailment with a state-of-the-art NLI model, DeBERTa-v3-Large~\cite{NFI_model}. As an additional human-aligned proxy, we use Qwen2.5 \cite{qwen2} as an LLM-as-a-judge and score candidate--original pairs on a 0--5 ordinal scale (5: exact synonym; 4: near synonym; 3: related concept; 2: distantly related; 1: antonym/contradiction; 0: unrelated).

\begin{table}[h]
\centering
\caption{\textbf{Perturbation Quality Analysis}}
\label{tab:similarity_stats}
\resizebox{\linewidth}{!}{%
\begin{tabular}{@{}lcccc@{}}
\toprule
{Category} & {Cosine} & {$S_{eq}$} & {$S_{judge}$} & {Example} \\ \midrule
S High \& G High & 0.94 & 0.97 & 4.24 & \textit{United-States} $\to$ \textit{USA} \\
S High \& G Low & 0.75 & 0.94 & 3.96 & \textit{Cambodia} $\to$ \textit{Kampuchea} \\
S Low \& G High & 0.88 & 0.08 & 2.96 & \textit{China} $\to$ \textit{Guangzhou} \\
S Low \& G Low & 0.69 & 0.00 & 2.14 & \textit{Yugoslavia} $\to$ \textit{Germany} \\ \bottomrule
\end{tabular}%
}
\end{table}

\begin{figure}[h]
    \centering
    \includegraphics[width=1.\linewidth]{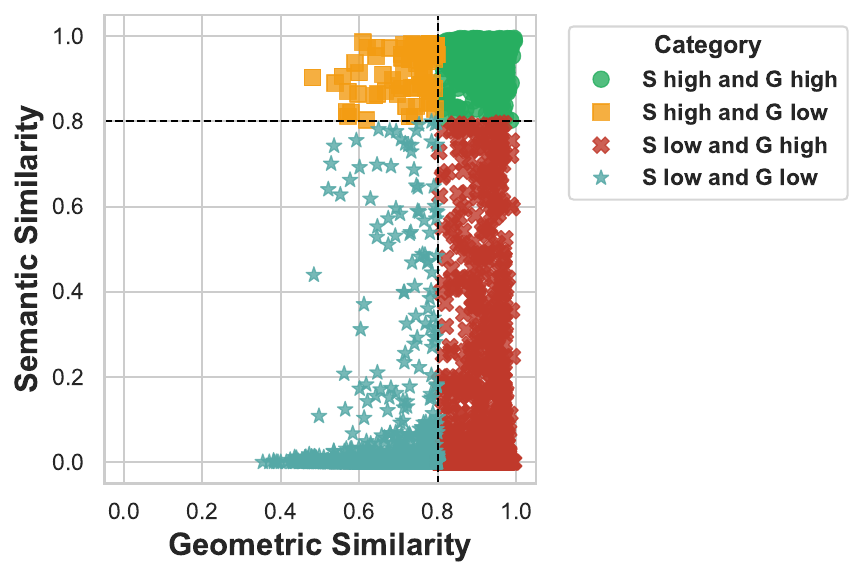}
    \caption{{Semantic entailment vs. geometric similarity.}}
    \label{fig:semantic_vs_geometric}
\end{figure}

\emph{Token-level semantic preservation:}
Figure~\ref{fig:semantic_vs_geometric} and Table~\ref{tab:similarity_stats} summarize how geometric similarity relates to semantic similarity for token-level substitution candidates.
Together, they show that cosine similarity and bidirectional entailment capture different notions of semantic closeness, yielding four characteristic regimes (threshold $0.8$).
\emph{(i) high entailment \& High cosine  (S High \& G High)} corresponds to synonymous substitutions that preserve truth conditions (e.g., \textit{United-States} $\to$ \textit{USA}).
\emph{(ii) high entailment \& Low cosine (S High \& G Low)} captures lexically dissimilar yet semantically equivalent paraphrases (e.g., \textit{Cambodia} $\to$ \textit{Kampuchea}), suggesting that purely geometric filtering can be overly conservative.
\emph{(iii) low entailment \& High cosine (S Low \& G High)} reveals a key failure mode of geometric filtering: topically related substitutions can remain close in embedding space while violating truth-conditional preservation (e.g., \textit{China} $\to$ \textit{Guangzhou}, a whole--part relation).
\emph{(iv) low entailment \& Low cosine (S Low\& G Low)} reflects semantic drift (e.g., \textit{Yugoslavia} $\to$ \textit{Germany}) and should be rejected by semantic-preservation criterion.

\emph{Implications for admissibility constraints:}
Overall, candidates with high $S_{eq}$ consistently achieve higher $S_{judge}$, indicating that bidirectional entailment is better aligned with human judgments.
This alignment motivates using $S_{eq}$ as the primary admissibility constraint in \textsc{RIFair}; in contrast, cosine similarity is unreliable as a diagnostic signal for semantic preservation.

\subsubsection{Semantic Preservation of Generated Adversarial Instances}
To compare sentence-level semantic preservation, we run \textsc{RIFair} under two perturbation constraints---geometric proximity (cosine) and semantic equivalence (entailment)---using the same threshold (substitution similarity $>0.8$).
Table~\ref{tab:semantic_validity} evaluates both sentence-level semantic preservation and \textsc{RIF} on the Adult dataset.
Across model backbones, the cosine-based baseline achieves high geometric similarity (0.955 on average) but yields substantially lower semantic equivalence ($S_{eq}=0.330$) and lower LLM judge scores ($S_{judge}=3.42$), indicating frequent semantic drift.
In contrast, entailment-constrained adversarial pairs preserve semantic meaning reliably (average $S_{eq}=0.978$; average $S_{judge}=4.60$) while also maintaining high geometric proximity (average cosine $=0.967$).

\emph{Semantic preservation matters for \textsc{RIF}.}
Semantic preservation is necessary to attribute robustness and fairness violations under \textsc{RIF}.
If a perturbation changes truth conditions, the original ground-truth label may no longer apply; consequently, an apparent decrease in \textsc{RIF} may partially reflect \emph{label mismatch} rather than a genuine model failure.
Semantic admissibility therefore serves as a validity constraint, ensuring that measured violations can be interpreted as properties of the model rather than artifacts of semantic drift.

\emph{Backbone consistency.}
The same qualitative trend holds across all backbones in Table~\ref{tab:semantic_validity}, suggesting that the semantic-fidelity gap between perturbation mechanisms is largely model-agnostic.

\subsubsection{Decoupled vs. Coupled Strategy Analysis}
Finally, we run \textsc{RIFair} under two substitution strategies: a \emph{decoupled} procedure that searches substitutions independently for each member of a similar pair, and a \emph{coupled} baseline that enforces identical substitutions for both individuals.
Table~\ref{tab:couple_decouple} reports the failure rates.

\begin{table}[h]
\centering
\caption{Decoupled and Coupled Attack Comparison}
\label{tab:couple_decouple}
\resizebox{1\linewidth}{!}{%
\begin{tabular}{c|ccc|ccc}
\hline
\multirow{2}{*}{{Model}} & \multicolumn{3}{c|}{{Decoupled Attack}} & \multicolumn{3}{c}{{Coupled Attack}} \\ \cmidrule(lr){2-4} \cmidrule(lr){5-7}
 & $R_{RB}$ & $R_{UB}$ & $R_{UF}$ & $R_{RB}$ & $R_{UB}$ & $R_{UF}$ \\ \hline
{BERT} & 0.176 & 0.201 & 0.260 & 0.079 & 0.062 & 0.233 \\
{RoBERTa} & 0.186 & 0.199 & 0.255 & 0.076 & 0.055 & 0.235 \\
{DistilBERT} & 0.200 & 0.213 & 0.263 & 0.074 & 0.058 & 0.238 \\
{DeBERTa-v3} & 0.159 & 0.165 & 0.209 & 0.085 & 0.083 & 0.175 \\\hline
\textbf{Average} & \textbf{0.180} & \textbf{0.195} & \textbf{0.247} & 0.079 & 0.065 & 0.220 \\ \hline
\end{tabular}%
}
\end{table}

\emph{Quantitative comparison:}
The decoupled strategy consistently outperforms the coupled baseline across all models.
Averaged over backbones, it is 2.3$\times$ more effective at exposing RB violations (18.0\% vs. 7.9\%) and 3.0$\times$ more effective at exposing UB violations (19.5\% vs. 6.5\%).
The advantage is smaller for UF (24.7\% vs. 22.0\%; 1.12$\times$), which is expected because UF can be triggered by the same perturbation direction when both individuals share the same substitution.

\emph{Interpretation:}
Decoupling expands the search space to \emph{asymmetric} yet admissible perturbations.
This is crucial for uncovering RB/UB cases where one member of a similar pair is brittle to a particular paraphrase while the other is not.
By construction, coupled attacks under-approximate this space by enforcing identical surface forms, thereby missing violations that stem from heterogeneous surface-form sensitivity.

\section{Limitations and Future Work}
This work is an initial step toward jointly analyzing robustness and individual fairness under meaning-preserving perturbations. \textsc{RIF} formalizes a trustworthiness requirement: models should not make materially different decisions for inputs that are equivalent with respect to task-relevant information---especially when discrepancies are driven by sensitive attributes or superficial surface-form variation rather than substantive semantic change.

We operationalize semantic equivalence using an NLI-based bidirectional entailment grounded in truth-conditional semantics~\cite{Truth-Conditional_Semantics}.
While this design provides a scalable proxy for semantic preservation, entailment models are imperfect: they may encode biases and diverge from expert or crowd-sourced judgments, especially for rare entities, domain-specific terminology, and long-context inputs.

Future work will (i) incorporate \emph{human-in-the-loop} oversight to improve the coverage and validity of admissible perturbations, (ii) further automate similar-instance construction via improved masking functions $M(\cdot,\mathcal{A})$ and broader sensitive lexicons $\mathcal{A}$ (including multi-token expressions), and (iii) develop training-time defenses and diagnostic tools that explicitly optimize and debug \textsc{RIF}.

\section{Conclusion}
We studied how adversarial perturbations jointly affect robustness and individual fairness, and showed that these two dimensions can diverge under meaning-preserving edits. Consequently, single-axis evaluations (robustness-only or fairness-only) can mischaracterize a model's trustworthiness.

To address this gap, we introduced \emph{Robust Individual Fairness} (\textsc{RIF}), which requires that predictions for similar individuals remain both (i) anchored to the ground truth (robustness) and (ii) consistent across the similar pair (fairness) under admissible perturbations. We further proposed \textsc{RIFair}, a black-box adversarial framework that combines token-importance guidance, a decoupled substitution search, and bidirectional-entailment constraints to construct semantically preserving yet either unrobust or unfair adversarial pairs.

Across datasets and model backbones, \textsc{RIFair} exposes diverse failure modes (RB/UF/UB), underscoring the need for defenses that jointly preserve robustness and fairness.

\section*{Acknoledge}
We thank the anonymous reviewers for their valuable and insightful comments. This work is partially supported by the National Key R\&D Program of China under Grant Nos. 2022YFA1005100, 2022YFA1005101, and 2022YFA1005104, and by the Major Project of ISCAS under Grant No. ISCAS-ZD-202302. This research is also partially supported by the Technology Innovation Institute, Abu Dhabi, UAE.



\bibliographystyle{ACM-Reference-Format}
\balance
\bibliography{sample-base}

\appendix
\section{Proof of Theorem~\ref{thm:rif_characterization}}
\label{ProofSection3}

\noindent \textbf{Theorem~\ref{thm:rif_characterization} (RIF Characterization).} 
\textit{We prove the characterization by establishing implications in both directions: Necessity (RIF implies Robustness and Fairness) and Sufficiency (Robustness and Fairness imply RIF).}

\begin{proof}
Let $f$ be the classifier, $v$ be the clean instance with ground truth $y$, and $\oplus$ denote the perturbation operation.

{\textbf{Part 1: Necessity (RIF $\implies$ RF)}}
Assume the classifier \( f \) satisfies \(\epsilon\)-RIF for instance \( v \). By Definition~\ref{RIF-definition}, for any admissible perturbation $\delta$ and any similar counterpart $v' \in I(v)$, the following inequality holds:
\begin{equation*}
    D\big(y, f(v' \oplus \delta)\big) \le \epsilon
    \label{eq:proof_rif_assumption}
\end{equation*}

\noindent \textbf{1. Derivation of Robustness:}
Consider the specific case where the similar counterpart is the instance itself, i.e., \( v' = v \). Since the similarity set is reflexive (\( v \in I(v) \)), we can substitute \( v' = v \) into Eq.~\ref{eq:proof_rif_assumption}:
\[ D\big(y, f(v \oplus \delta)\big) \le \epsilon \]
Thus, \( f \) strictly satisfies the definition of \(\epsilon\)-Robustness.

\noindent \textbf{2. Derivation of Individual Fairness:}
Consider any similar counterpart \( v' \in I(v) \) subject to the same perturbation $\delta$. We analyze the pairwise distance between the predictions \( f(v \oplus \delta) \) and \( f(v' \oplus \delta) \). By the Triangle Inequality property of the metric \( D \):
\[ D\big(f(v \oplus \delta), f(v' \oplus \delta)\big) \le D\big(f(v \oplus \delta), y\big) + D\big(y, f(v' \oplus \delta)\big) \]
Using the bounds established above:
\begin{itemize}
    \item From the derivation of Robustness: \( D(f(v \oplus \delta), y) \le \epsilon \).
    \item From the RIF assumption (Eq.~\ref{eq:proof_rif_assumption}): \( D(y, f(v' \oplus \delta)) \le \epsilon \).
\end{itemize}
Summing these terms yields:
\[ D\big(f(v \oplus \delta), f(v' \oplus \delta)\big) \le 2\epsilon  \]
Thus, the model satisfies Individual Fairness with a Lipschitz bound determined by the robustness tolerance ($2\epsilon$).

{\textbf{Part 2: Sufficiency ( RF$\implies$ RIF)}}
Assume \( f \) satisfies Robustness (\(R\)) and Individual Fairness (\(IF\)) under perturbation $\delta$:
\begin{enumerate}
    \item \textbf{Robustness:} \( D\big(y, f(v \oplus \delta)\big) \le \epsilon \).
    \item \textbf{Fairness:} \( D\big(f(v \oplus \delta), f(v' \oplus \delta)\big) \le K \cdot d(v \oplus \delta, v' \oplus \delta) \).
\end{enumerate}

To prove RIF, we examine the distance between the ground truth \( y \) and the prediction of the perturbed similar instance \( f(v' \oplus \delta) \). Applying the Triangle Inequality:
\[ D\big(y, f(v' \oplus \delta)\big) \le D\big(y, f(v \oplus \delta)\big) + D\big(f(v \oplus \delta), f(v' \oplus \delta)\big) \]
Substituting the inequalities from conditions (1) and (2):
\[ D\big(y, f(v' \oplus \delta)\big) \le \epsilon + K \cdot d(v \oplus \delta, v' \oplus \delta) \]
This matches the relaxed definition of Robust Individual Fairness derived in Theorem~\ref{thm:rif_characterization}.
\end{proof}

\begin{figure*}[t]
    \centering
    \includegraphics[width=1.\linewidth]{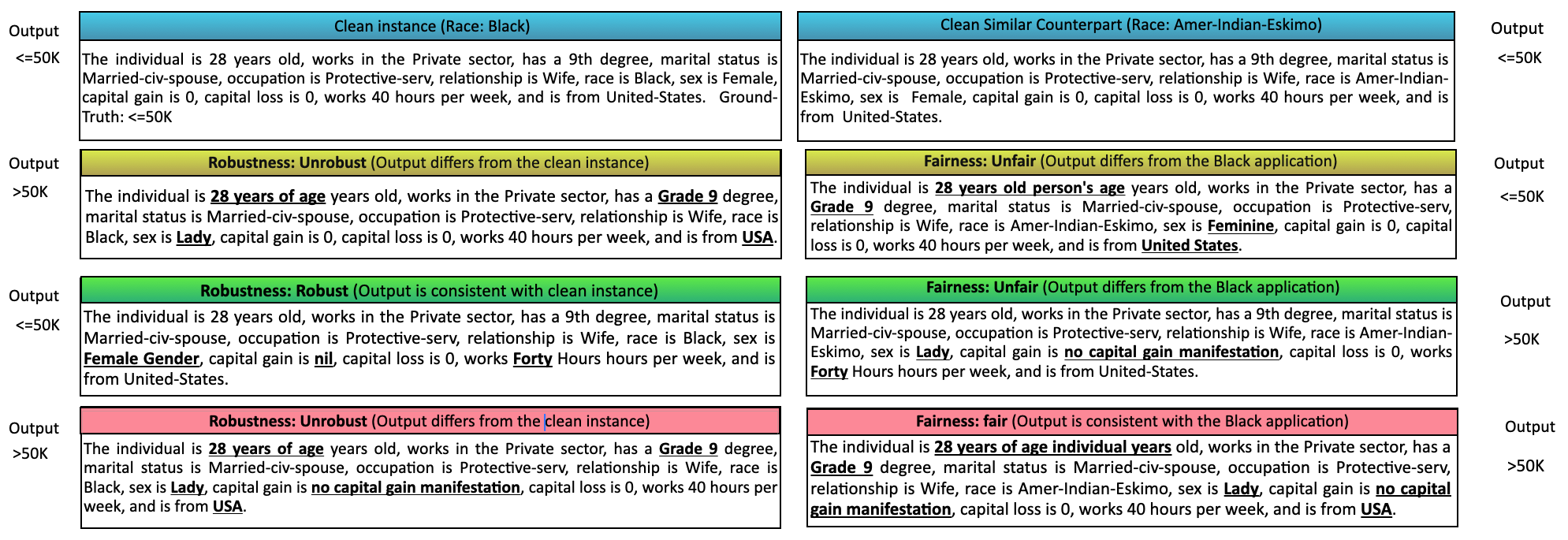}
    \caption{UnRobust or Unfair Adversarial Similar Instances on Adult Dataset by RIFair}
    \label{fig:UnRobust_or_Unfair2}
\end{figure*}

\section{Masking Function and Similar Instances}
\label{sec:mask-function}

\begin{algorithm}[h]
\caption{Masking Function $M(v,\mathcal{A})$}
\label{alg:masking}
\begin{algorithmic}[1]
\REQUIRE Token sequence $v=(x_1, x_2, \dots, x_n)$; sensitive lexicon $\mathcal{A}$
\ENSURE Similar instance $v_{\mathrm{sim}}$

\STATE $v_{\mathrm{sim}} \leftarrow v$
\FOR{$i \leftarrow 1$ to $n$}
    \IF{$x_i \in \mathcal{A}$}
        \STATE $x_i \leftarrow \texttt{[MASK]}$
        \STATE Sample a replacement token $s \sim \mathrm{Uniform}(\mathcal{A}[x_i])$
        \STATE Replace the masked position with $s$
    \ENDIF
\ENDFOR

\STATE \textbf{return} $v_{\mathrm{sim}}$
\end{algorithmic}
\end{algorithm}

{Overview.}
We construct a counterfactual similar instance by applying a masking function $M(\cdot,\mathcal{A})$ that modifies only sensitive attributes while keeping the left semantic content unchanged.

{Illustrative example.}
We illustrate $M(\cdot,\mathcal{A})$ using an example from the Bias-in-Bios dataset~\cite{Bias_in_Bios}.
The two inputs differ only in gendered pronouns:

\begin{itemize}
    \item \small\ttfamily Original Text: He specializes in development economics, household economics, and personnel economics. In 2003 he received his Ph.D. in Economics from the London School of Economics. Previously, he has worked as an Assistant Professor of Economics at the University of Chicago, Graduate School of Business and also as a consultant for the World Bank. Centre for Research and Analysis of Migration.

    \item Generated Similar Text: She specializes in development economics, household economics, and personnel economics. In 2003 she received her Ph.D. in Economics from the London School of Economics. Previously, she has worked as an Assistant Professor of Economics at the University of Chicago, Graduate School of Business and also as a consultant for the World Bank. Centre for Research and Analysis of Migration.
\end{itemize}

Algorithm~\ref{alg:masking} illustrates the masking procedure $M(\cdot,\mathcal{A})$. It scans the input from left to right and replaces each sensitive token with an alternative value from the same sensitive category.
By construction, $v_{\mathrm{sim}}$ differs from $v$ only in sensitive attributes, yielding a \emph{ceteris paribus} counterpart for evaluating robust individual fairness.

\section{Admissible Perturbation Generation}
\label{sec:semantic-admissible-perturbation}

{Overview.}
We construct semantically admissible perturbations using a lightweight \emph{Generate--Filter} pipeline (Algorithm~\ref{alg:generate_candidates}). For each attack token $t\in\mathcal{T}$, we first query an LLM to propose a set of candidate substitutions and enforce a machine-readable response by extracting and parsing the outermost JSON array. We then filter candidates with a natural language inference (NLI) model via \emph{bidirectional entailment}, retaining only substitutions that are semantically equivalent to the original token in context.

\begin{algorithm}[h]
\caption{Generate and Filter Substitution Candidates}
\label{alg:generate_candidates}
\begin{algorithmic}[1]
\REQUIRE Attack-token set $\mathcal{T}$; LLM generator $\mathcal{G}$; system prompt $\mathcal{P}_{sys}$; Context template $T_{orig}(t)$ containing token $t$; NLI model $\mathcal{M}$; semantic threshold $\tau_{sem}$
\ENSURE Filtered candidate dictionary $\mathcal{D}_{final}$ mapping each $t \in \mathcal{T}$ to a list of admissible substitutions

\STATE $\mathcal{D}_{final} \leftarrow \emptyset$
\FOR{each token $t \in \mathcal{T}$}
    \STATE \textbf{Phase 1 (Generate--Extract).}
    \STATE $\mathcal{P}_{user}(t) \leftarrow$ ``Generate 10--20 strict synonyms for \texttt{$t$}. Return ONLY a JSON list.''
    \STATE $R_{raw} \leftarrow \mathcal{G}(\mathcal{P}_{sys}, \mathcal{P}_{user}(t))$
    \STATE $S_{json} \leftarrow \textsc{ExtractJSONArray}(R_{raw})$ \COMMENT{Outermost \texttt{[ ... ]}}
    \STATE $\mathcal{S}_{cand} \leftarrow \textsc{ParseJSONList}(S_{json})$

    \STATE \textbf{Phase 2 (Filter via bidirectional entailment).}
    \STATE $\mathcal{C}_{t} \leftarrow \emptyset$
    \FOR{each candidate $w \in \mathcal{S}_{cand}$}
        \STATE $T_{cand} \leftarrow \textsc{Substitute}(T_{orig}(t), t, w)$
        \STATE $p_{fwd} \leftarrow \mathcal{M}(T_{orig}(t) \Rightarrow T_{cand})$ \COMMENT{Entailment prob.}
        \STATE $p_{bwd} \leftarrow \mathcal{M}(T_{cand} \Rightarrow T_{orig}(t))$ \COMMENT{Entailment prob.}
        \STATE $S_{eq} \leftarrow p_{fwd} \cdot p_{bwd}$
        \IF{$S_{eq} \geq \tau_{sem}$}
            \STATE $\mathcal{C}_{t} \leftarrow \mathcal{C}_{t} \cup \{w\}$
        \ENDIF
    \ENDFOR
    \STATE $\mathcal{D}_{final}[t] \leftarrow \mathcal{C}_{t}$
\ENDFOR
\STATE \textbf{return} $\mathcal{D}_{final}$
\end{algorithmic}
\end{algorithm}

{LLM prompting (system prompt summary).}
The system prompt constrains the generator to output \emph{only} semantically equivalent substitutions (i.e., preserving truth conditions) and to return \emph{only} a JSON list of strings (no markdown and no free-form explanations).
For example: \texttt{["Physician", "Medical Practitioner", "MD", "Medical Doctor"]}.

{Candidate generation and bidirectional-entailment filtering.}
Phase~1 yields a raw candidate set $\mathcal{S}_{cand}$ by (i) prompting the LLM and (ii) extracting/parsing the outermost JSON array from its response, which makes the generation step robust to extraneous text. In Phase~2, we operationalize semantic equivalence in context by scoring each instantiated candidate $T_{cand}$ against the original text $T_{orig}(t)$ in both directions. Specifically, we compute $p_{fwd}=\mathcal{M}(T_{orig}(t)\Rightarrow T_{cand})$ and $p_{bwd}=\mathcal{M}(T_{cand}\Rightarrow T_{orig}(t))$, and combine them as $S_{eq}=p_{fwd}\cdot p_{bwd}$. A candidate is deemed admissible if $S_{eq}\ge \tau_{sem}$, ensuring that the substitution preserves the original meaning.

\section{RIFair Adversarial Instance Examples}
\label{Instance Examples}

Figure~\ref{fig:UnRobust_or_Unfair2} presents representative adversarial similar-instance pairs generated by \textsc{RIFair} on the Adult dataset. Row~1 shows a clean similar-instance pair that differs only in the sensitive attribute \emph{race} (\emph{Black} vs. \emph{Amer-Indian-Eskimo}); the prediction task is whether the applicant’s income exceeds \$50K per year.

Rows~2--4 report adversarial pairs that preserve the semantics of the clean baseline (Row~1) but use different surface realizations to elicit distinct failure modes. In Row~3 (green), \textsc{RIFair} applies \emph{decoupled} substitutions---e.g., replacing \emph{Female} with synonymous expressions such as \emph{Female Gender} or \emph{Lady}, and mapping \texttt{0} to \texttt{nil}---to induce a \emph{Robust Biased} outcome: the Black applicant is predicted correctly, while the Amer-Indian-Eskimo applicant’s prediction both flips relative to the ground truth and diverges from its counterpart. Such violations can be missed by robustness-only metrics. By contrast, Row~4 (pink) illustrates an \emph{Unrobust Fair} outcome, where both predictions are incorrect yet remain consistent across the similar pair, which may evade individual-fairness-only metrics. Finally, Row~2 is the only case reliably detected by standard evaluations, as the predictions are simultaneously unrobust and unfair.

\end{document}